\documentclass[10pt,twocolumn]{article}
\usepackage[a4paper,margin=0.75in]{geometry}
\usepackage{graphicx}
\usepackage{amsmath,amssymb}
\usepackage{hyperref}
\usepackage{titlesec}
\usepackage{caption}
\usepackage{multicol}
\usepackage{url}
\usepackage{xcolor}
\usepackage{orcidlink}

\hypersetup{
    colorlinks = true, 
    linkcolor=[rgb]{0,0.0,0.5},
    citecolor=[rgb]{0,0.0,0.5},
    urlcolor=[rgb]{0,0.0,0.5},
    anchorcolor = blue,
}

\usepackage[doi=true,url=true,isbn=false,maxbibnames=6,giveninits=true,style=numeric-comp,backend=biber,date=year,sorting=none]{biblatex}

\titleformat{\section}{\bfseries\large}{\thesection.}{0.5em}{}
\titleformat{\subsection}{\bfseries}{\thesubsection}{0.5em}{}

\title{\vspace{-1em}\textbf{LLM-Supported Formal Knowledge Representation for Enhancing Control Engineering Content with an Interactive Semantic Layer}}
\author{Julius Fiedler${}^*$\orcidlink{0009-0009-0163-9600}, Carsten Knoll${}^+$\orcidlink{0000-0001-7462-771X}, Klaus Röbenack${}^*$\orcidlink{0000-0002-3347-0864}\\
\small ${}^*$ Institute of Control Theory, TU Dresden\\
\small ${}^+$ Chair of Fundamentals of Electrical Engineering, TU Dresden\\
\small \texttt{julius.fiedler@tu-dresden.de}}
\date{GMA-FA 2.13, September 17, 2025}

\begin{document}
\twocolumn[
\maketitle
\vspace{-1em}
\begin{abstract}
The rapid growth of research output in control engineering calls for new approaches to structure and formalize domain knowledge. This paper briefly describes an LLM-supported method for semi-automated generation of formal knowledge representations that combine human readability with machine interpretability and increased expressiveness. Based on the \textit{Imperative Representation of Knowledge} (PyIRK) framework, we demonstrate how language models can assist in transforming natural-language descriptions and mathematical definitions (available as \LaTeX{} source code) into a formalized knowledge graph. As a first application we present the generation of an ``interactive semantic layer'' to enhance the source documents in order to facilitate knowledge transfer. From our perspective this contributes to the vision of easily accessible, collaborative, and verifiable knowledge bases for the control engineering domain.
\end{abstract}
\vspace{1em}
]

\section{Introduction}
Control engineering is a versatile field that includes a very broad spectrum of methods such as linear PID controllers, sliding mode, backstepping and nonlinear observers. Its great variety of application domains ranges from automotive and process systems to robotics and building automation and continuously expands. As new theories, algorithms, and practical insights emerge, the field faces a growing challenge: How to manage and transfer knowledge across its sub-disciplines and between theoretical research and practical applications.

Traditional publications (PDF files), while rich in content, are often difficult to access for non-experts in the domain, e.g. mechanical engineers developing a new machine. Digital tools are of limited help because knowledge is embedded in an interlaced combination of natural language, equations, and diagrams, which is difficult to query precisely and integrate computationally. To overcome this limitation, formal knowledge representation based on semantic technologies is a promising pathway.

\smallskip

This paper has two contributions: First, we present an approach to employing a large language model (LLM) for the semi-automatic extraction of knowledge from source documents and the subsequent enrichment of an existing base knowledge graph. Secondly, we use this knowledge graph to enhance the original source document with an ``interactive semantic layer'' containing additional information which alleviates the understanding of the content by human readers.

Note that this brief paper is based on a presentation given at the joint Workshop of GMA expert committees\footnote{See \url{https://www.uni-ulm.de/in/gma-fa-2-14/} for details.} 2.13 and 2.14 in Anif (Austria) on 2025-09-17. A more detailed paper is being prepared.

\section{Formal Knowledge Representation and the PyIRK Framework}
The central idea of knowledge representation is to express domain knowledge in a structured, explicit, and machine-readable form. \textit{Knowledge graphs} or \textit{ontologies} (which throughout this paper are used synonymously) provide a formal backbone for representing entities, their properties, and relationships. Following the definition of Studer \textit{et al.}~\cite{studer1998knowledge}, an ontology is a “formal, explicit specification of a shared conceptualization.” In this paradigm, knowledge is encoded as subject–predicate–object (SPO) triples that collectively form a semantic graph.

For engineering applications, such formalization enables advanced querying, consistency checking, and the integration of simulation and design data. However, creating and maintaining these ontologies is labor-intensive and requires expertise in both the application domain and knowledge engineering.

Furthermore, existing ontology languages such as the Web Ontology Language (OWL) are well established but not expressive enough\footnote{In fact the expressiveness of OWL is deliberately limited in order to guarantee decidability and efficient automated reasoning.} to capture the complex meaning structures which are key to understanding and applying control engineering concepts and methods. In particular, it is essential to precisely model the content of mathematical definitions and theorems which requires higher order logic, i.e. statements about statements.

On the other hand, the goal is to make formal knowledge representation accessible for engineers without requiring extensive dedicated training. Therefore, we developed the \textit{Imperative Representation of Knowledge} (PyIRK) framework \cite{knoll_fiedler_pyirk,knoll2024Imperative,knoll2025mocast}. Its main idea is to use the widespread and versatile programming language Python itself to express knowledge artifacts (therefore ``imperative'' knowledge representation), instead of a declarative language like XML or YAML which then would have to be parsed by a program. By leveraging the feature set of a dynamic programming language such as function creation at runtime and context managers, PyIRK allows to express complex structures like the setup-premise-assertion-compound of a mathematical theorem.

\smallskip

The core concepts of PyIRK are \textit{items}, \textit{relations}, and \textit{statements}. Items are used as nodes of the knowledge graph, statements represent node-edge-node subgraphs (also interpretable as semantic triple of \textit{subject}, \textit{predicate}, and \textit{object}), and relations specify the type of connection between the subgraph nodes. Additionally, there are \textit{literals} (strings or numbers) that can also serve as a target node (i.e. object).

Each node and edge in PyIRK's internal graph is associated with a globally unique resource identifier (URI) and a human-readable label, ensuring both traceability and interpretability. The framework supports direct interaction with the ontology through Python objects and also an interface for the semantic query language SPARQL.

%

\section{LLM-supported Knowledge Formali\-zation}

While the expressive power of PyIRK allows the formalization of complex relationships and its Python-based syntax eliminates the need to learn an ``exotic'' declarative language, it is still substantial effort necessary to convert a significant amount of knowledge (e.g. a chapter of a book) into a knowledge graph.

Therefore, in order to improve efficiency, we propose a semi-automated process in which LLMs assist as structured text processors. Naively, the LLM could be used to generate the required python code directly from the source material\footnote{The question why a knowledge graph is necessary at all in contrast to ``just using'' an LLM as question-answering agent is discussed in Sect. \ref{sec_discussion}.}. In practice, this does not produce good results because the PyIRK specification is likely not part of the training data of public LLMs.

We therefore employ a two-step process, first converting the source into a markdown-like format called \textit{Formal Natural Language} (FNL) and afterwards utilizing this intermediate result to algorithmically construct the PyIRK code. Thereby, FNL is specified in each prompt as a heavily simplified version of English with a controlled vocabulary and strict structure of subject, predicate, object. Additionally, scoping information, e.g., to represent the premise of a theorem, is encoded in nested bullet point lists. While our FNL definition also has not been part of the LLM training data, it is siginficantly easier to specify it in the prompt (see part 1 below), than PyIRK.

\smallskip

As source documents, LaTeX source code is particularly suitable for our approach, since mathematical expressions and other special characters, which are omnipresent in the control engineering domain, are represented with unambiguous syntax. While most scientific knowledge in our domain is first published in the form of papers -- typically with $<25$ pages and a rather narrow focus -- our approach is best suited to be applied to books, and more precisely monographs\footnote{This restriction excludes books which are basically a collection of papers.}. This is because such books use consistent notation and language conventions over hundreds of pages and thus prevent the necessity of identifying and aligning symbols and words across multiple sources.

Before the actual conversion takes place, a preprocessing step is necessary: Snippet-delimiter comments are added into the LaTeX source code to subdivide the whole document into small units, typically containing one to five sentences (depending on complexity).

For each such snippet an LLM-prompt is generated based on an extensive template.
This template\footnote{See file \texttt{data/templates/prompt01\_template\_german.md} in the source repository \cite{Fiedler2025stafo_repo}} consists of $\approx 11$\,KB (240 lines) of detailed instructions and examples formulated in markdown syntax and is structured into parts by the following headlines:

\begin{enumerate}
 \item Description of allowed formalized statements (170 lines)
 \item Remarks and instructions
 \item The LaTeX source code which was already processed
 \item The formalized statements which where extracted from that LaTeX source code
 \item The new LaTeX source code from which you should generate new formalized statements
 \item LaTeX source code which follows the previous snippet
 \item Final Instructions
\end{enumerate}

In this template parts 3 to 6 are extended with the appropriate strings from the LaTeX source document and the formalized results obtained so far. The resulting prompt string is then sent to a public commercial LLM service\footnote{We use Google Gemini, but other LLMs are expected to deliver similar results.} which responds with a string of FNL statements (step 1a in Fig.~\ref{fig_workflow1}).

The next step (1b) is to manually review these statements and perform amendments if necessary. While this step is time-consuming, it is still much faster than manually writing FNL or PyIRK statements. From our experience about 10\% to 20\% of the FNL statements (depending on the complexity of the respective snippet) need manual intervention. Obviously, step 1b is the bottleneck in the current approach but we are optimistic to further reduce the necessity of manual intervention in future versions (see Sec. \ref{sec_outlook})


Finally, the resulting revised FNL code can be inserted in the prompt template to process the next LaTeX snippet or be converted by ``ordinary'' (i.e., non-LLM-based) algorithms to obtain a knowledge graph represented in PyIRK code (step 2 in Fig.~\ref{fig_workflow1}).

\section{Application: Interactive Semantic Docu\-ment Layer}

Once the knowledge is formalized, applications can be built on top of the knowledge graph. The mid-term goal of the authors is to build an interactive assistant which can help answer control theoretic questions and design control facilities. However, for such a system to be useful in nontrivial use cases, a ``critical mass'' of consistently formalized knowledge is required, which is not yet available.

To demonstrate the usefulness of formal knowledge representation
already with a much smaller knowledge graph, we propose to inject suitable statements directly back to the source document in the form of an ``interactive semantic layer''. This layer initially is invisible for the reader to allow for undistracted reading. However, the reader has the option to interactively unhide additional information (based on the knowledge graph) to make the meaning of certain words or symbols more precise.

The rationale for this approach is the following:
Scientific texts are usually written under the assumption that the readers linearly read, understand and memorize the work from beginning to the end. For example, if a concept or a notation is introduced on page $n$ it is assumed to be known on every following page. While this assumption is necessary to avoid redundancies, it makes it difficult for readers to concentrate on those parts which they deem to be relevant. The semantic layer offers additional explicit information which otherwise would have to be taken from the context. This a) reduces the necessity for tedious page flipping to find a particular definition and b) might dispel doubts in case of misunderstandings. Because a priori it is unknown which additional information will turn out to be useful, a strategy of ``hidden complete redundancy'' is used: introduced technical terms like \textit{orthocomplement} or symbolic notation like  $\mathbb{U}^\perp$ are explained on \textit{every} occasion but only displayed interactively upon user request.

\begin{figure}[h]
\centering
\includegraphics[width=0.99\columnwidth]{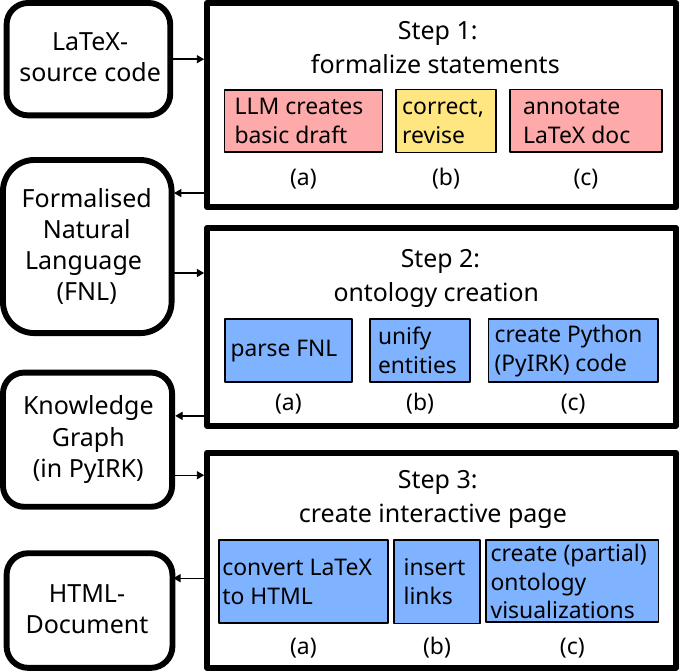}
\caption{Conversion workflow \cite{Fiedler2025stafo_repo}. Red: automated process with LLM support, yellow: manual process, blue: automated process.}
\label{fig_workflow1}
\end{figure}

On a technical level this interactivity is implemented by converting the original LaTeX source to HTML and add special elements (\texttt{div}-tags with \texttt{class="tooltip"}) which become visible only if the user hovers the respective words or symbols. To fill those tooltip-tags with content the LLM-based step 1c and the algorithmic steps 3a - 3c  are necessary (see Fig.~\ref{fig_workflow1}).


\begin{figure}[h]
\centering
\includegraphics[trim=0 0 45mm 0, clip, width=0.99\columnwidth]{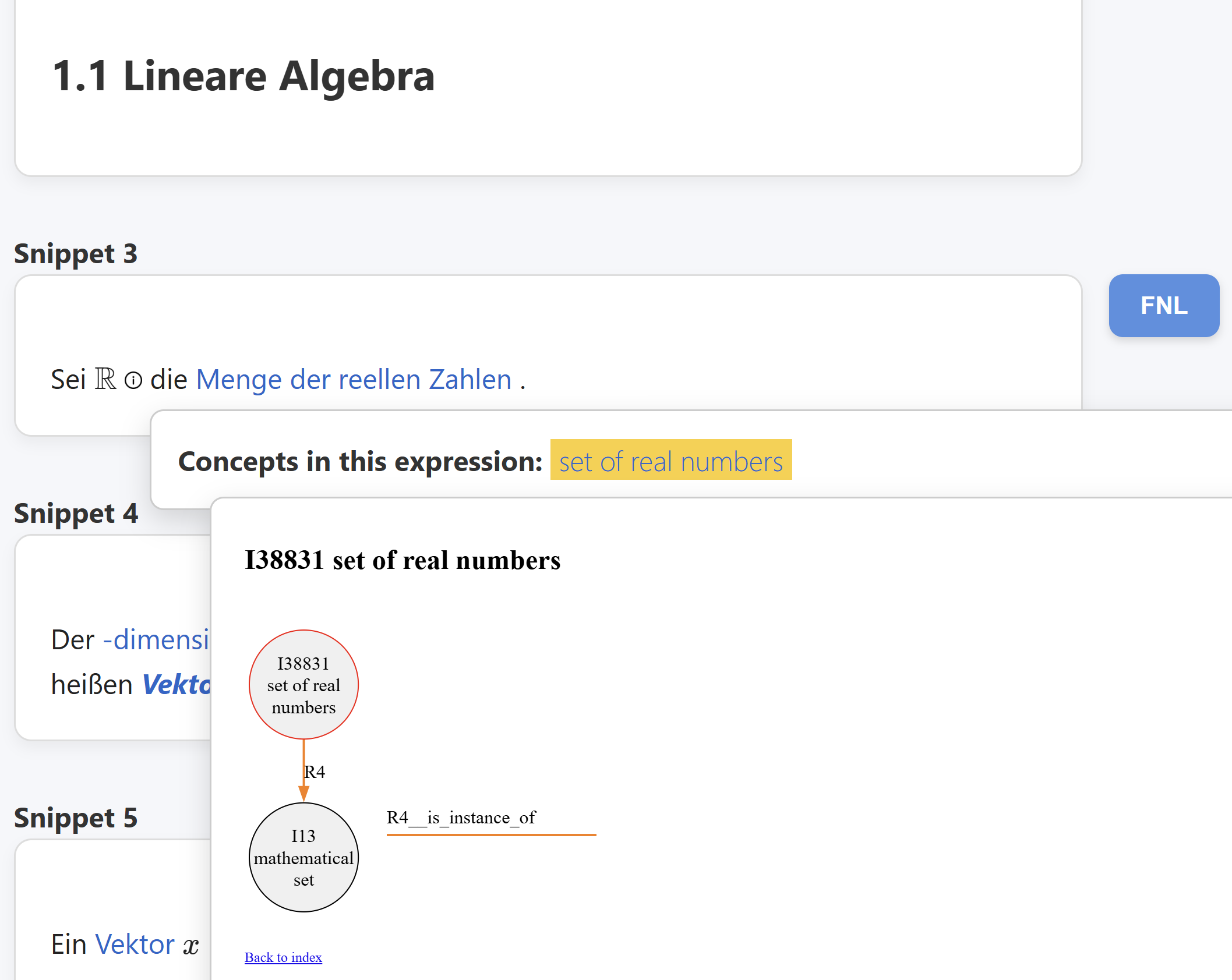}
\caption{Screenshot of the conversion result of snippets 2 to 5 of \cite{robenack2017nichtlineare}. Two tooltips are unhidden.}
\label{fig_screenshot1}
\end{figure}

\smallskip
To test this approach we applied the pipeline of all three steps to the first two sections (spanning eight  pages) of \cite{robenack2017nichtlineare} and obtained an HTML file with $\approx 700$ tooltip elements. As illustration, Fig.~\ref{fig_screenshot1} depicts the converted snippets 2 to 5 including two unhidden tooltip elements.



\section{Discussion}
\label{sec_discussion}

Given that the presented approach heavily relies on the LLMs capabilities, the question arises: Is the effort of knowledge formalization justified if one could simply ask the LLM directly?

From our perspective the creation of a knowledge graph has several advantages:
\begin{itemize}
 \item It offers an explicit and transparent knowledge base which can be referenced with high precision by using URIs of individual concepts and statements (unlike the knowledge implicitly stored inside the LLM weights).
 \item Due to its text-based representation, changes to the knowledge base can be traced by version control systems. This significantly facilitates collaborative maintenance.
 \item Semantic representation allows to integrate knowledge from different sources (e.g. different books) without redundancy.
 \item Extracting knowledge from a knowledge graph is computationally much more efficient than extracting the knowledge from an LLM with a suitable parameter number.
\end{itemize}

Note that in our approach we only use the LLM as an auxiliary tool in intermediate steps along with human supervision.

%

\section{Outlook}
\label{sec_outlook}

As written above, the presented approach still has step (1b) as a major bottleneck. In the future we aim to reduce the manual correction work significantly by inserting an LLM-based supervisor along with algorithmic quality assurance measures before the human corrector.

Another limitation of the current approach is its reliance on the availability of LaTeX source code of the source document. In a future version, we aim to enable the processing of PDF documents without losing quality. However, this comes with significant technical challenges.


Additionally, a user study to evaluate the knowledge transfer effects of the interactive semantic layer would be interesting.

The next big step then is to create an actual assistant that can answer control theoretic questions and explain the answer by referring to the appropriate URIs. Nevertheless, we think that the concept of interactive semantic layers has merit on its own.



\printbibliography

\end{document}